\def\year{2019}\relax
\documentclass[letterpaper]{article} 
\usepackage{aaai19}  
\usepackage{times}  
\usepackage{helvet}  
\usepackage{courier}  
\usepackage{url}  
\usepackage{graphicx}  
\usepackage{algorithm,algpseudocode}
\newtheorem{exmp}{Example}
\algnewcommand\algorithmicinput{\textbf{Input:}}
\algnewcommand\Input{\item[\algorithmicinput]}
\algnewcommand\algorithmicoutput{\textbf{Output:}}
\algnewcommand\Output{\item[\algorithmicoutput]}
\algnewcommand\algorithmicforeach{\textbf{for each}}
\algdef{S}[FOR]{ForEach}[1]{\algorithmicforeach\ #1\ \algorithmicdo}

\usepackage{amsmath}
\usepackage{subcaption}
\begin{document}
%
\title{A Clustering and Demotion Based Algorithm for\\ Inductive Learning of Default Theories}

\author{
    Huaduo Wang \and Farhad Shakerin\and Gopal Gupta\\
    Computer Science Department, The University of Texas at Dallas, Richardson, USA\\
    \{huaduo.wang,farhad.shakerin,gupta\}utdallas.edu
}

\maketitle
\begin{abstract}
We present a clustering- and \textit{demotion}-based algorithm called Kmeans-FOLD to induce \textit{nonmonotonic} logic programs from positive and negative examples. Our algorithm improves upon---and is inspired by---the FOLD algorithm. The FOLD algorithm itself is an improvement over the FOIL algorithm. Our algorithm generates a more concise logic program compared to the FOLD algorithm. Our algorithm uses the K-means-based clustering method to cluster the input positive samples before applying the FOLD algorithm. Positive examples that are covered by the partially learned program in intermediate steps are not discarded as in the FOLD algorithm, rather they are \textit{demoted}, i.e., their weights are reduced in subsequent iterations of the algorithm.  Our experiments on the UCI dataset show that a combination of K-Means clustering and our demotion strategy produces significant improvement for datasets with more than one cluster of positive examples.  The resulting induced program is also more concise and therefore easier to understand compared to the FOLD and ALEPH systems, two state of the art inductive logic programming (ILP) systems.  

\end{abstract}

\section{Introduction}
Dramatic success of machine learning has led to a torrent of Artificial Intelligence (AI) applications. However, the effectiveness of these systems is limited by the machines' current inability to explain their decisions and actions to human users. That is mainly because the statistical machine learning methods produce models that are complex algebraic solutions to optimization problems such as risk minimization or data likelihood maximization. Lack of intuitive descriptions makes it hard for users to understand and verify the underlying rules that govern the model. Also, these methods cannot produce a justification for a prediction they compute for a new data sample.

The Explainable AI program \cite{xai} aims to create a suite of machine learning techniques that: a) Produce more explainable models, while maintaining a high level of prediction accuracy. b) Enable human users to understand, appropriately trust, and effectively manage the emerging generation of artificially intelligent partners. 

Inductive Logic Programming (ILP) \cite{ilp} is one Machine Learning technique where the learned model is in the form of logic programming rules (Horn Clauses) that are comprehensible to humans. It allows the background knowledge to be incrementally extended without requiring the entire model to be re-learned. Meanwhile, the comprehensibility of symbolic rules makes it easier for users to understand and verify induced models and even edit/improve them. 

The ILP learning problem can be regarded as a search problem for a set of clauses learned from the training examples. The search is performed either top-down or bottom-up. A bottom-up approach builds most-specific clauses from the training examples and searches the hypothesis space by using generalization. This approach is not applicable to large-scale datasets, nor it can incorporate \textit{negation-as-failure} into the hypotheses. A survey of bottom-up ILP systems and their shortcomings can be found in \cite{sakama05}. In contrast, a top-down approach starts with the most general clauses and then specializes them. A top-down algorithm guided by heuristics is better suited for large-scale and/or noisy datasets \cite{quickfoil}.

The FOIL algorithm by Quinlan \cite{foil} is a popular top-down algorithm. FOIL uses heuristics from information theory called \textit{weighted information gain}. The use of a greedy heuristic allows FOIL to run much faster than bottom-up approaches and scale up much better. For instance, the QuickFOIL system \cite{quickfoil} can deal with millions of training examples in a reasonable time. However, scalability comes at the expense of losing accuracy if the algorithm is stuck in local optima and/or when the number of examples is insufficient. The former is an inherent problem in hill climbing search and the latter is due to the shrinking of examples during clause specialization. Also, elimination of already covered examples from the training set (to guarantee the termination of FOIL) causes a similar impact on the quality of heuristic search for the best clause. Therefore, the predicates picked-up by FOIL are not always globally optimal with respect to the concept being learned.

Based on our research, we believe that a successful ILP algorithm must satisfy the following criteria:

\begin{itemize}
\item It must employ heuristic-based search for clauses for the sake of scalability.
\item It should be able to figure out  relevant features, regardless of the number of current training examples.
\item It should be able to learn from incomplete data, as well as be able to distinguish between noise and exceptions.
\item It should generate concise and meaningful rules.
\end{itemize}

Unlike top-down ILP algorithms, statistical machine learning methods are bound to find the relevant features because they optimize an objective function with respect to global constraints. This results in models that are inherently complex and cannot explain what features account for a classification decision on any given data sample. 

The FOLD algorithm by Shakerin et al \cite{fold} is a new top-down algorithm inspired by the FOIL algorithm: its novelty is that it learns the exception predicates after first learning default rules. It does so by swapping the positive and negative examples, 
then recursively calling the algorithm. It generates predicates that capture exceptions unlike FOIL that learns negated predicate within the rules. The output of the FOLD algorithms is a set of default rules with exceptions coded in answer set programming \cite{cacm-asp,gelfond-kahl}. A sample that can be covered by any rule in the set would be classified as positive. Note that in any rule set that is learned, each rule merely covers a part of positive samples. A rule should avoid producing false positives since such wrongly classified samples cannot be eliminated by other rules. 

\begin{figure}
\begin{minipage}{.5\textwidth}
  \centering
    \begin{subfigure}[b]{0.5\linewidth}
        \centering
        \includegraphics[scale=0.33]{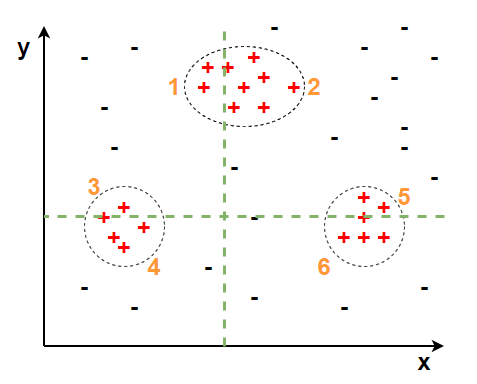}
        \caption{Sub-optimal literal selection}
        \label{fig:literalselectiona}
    \end{subfigure}\hfill
    \begin{subfigure}[b]{0.5\linewidth}
        \centering
        \includegraphics[scale=0.33]{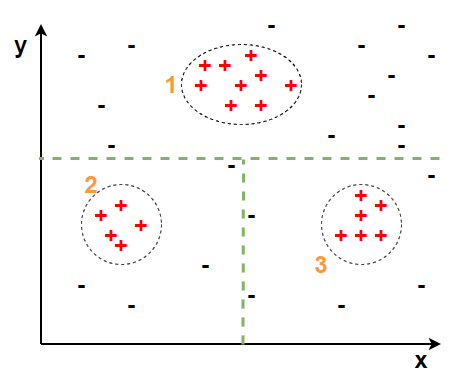}
        \caption{Optimal literal selection}
        \label{fig:literalselectionb}
    \end{subfigure}
    \caption{Literal selection}
    \label{fig:literalselection}
\end{minipage}
\end{figure}

Generally, the positive or negative data samples would scatter into several clusters in the hypothesis space of the dataset. Especially, for mid-size or large-size dataset, the positive or negative samples almost always scatter into more than one cluster. Logic programming/ASP is well suited to this scenario because rules are organized in a  disjunctive relation. The FOLD algorithm selects literals only based on information gain, which means hypothesis space would be divided along one dimension for each selection. The early stage of literal selection is very important: ideally, the algorithm should avoid dividing the hypothesis space by splitting up positive sample clusters (Figure \ref{fig:literalselectionb}), otherwise, it would lead to the problem of increased clusters in the hypothesis space and more rules will be needed to cover these positive sub-clusters (Figure \ref{fig:literalselectiona}). Often, more rules does not mean an increase in accuracy, rather, explainability is reduced due to significantly more rules for a human to understand. The search also becomes more complex requiring more computational power.

The idea of this paper is to apply ILP learning on each positive cluster to obtain a rule set, then produce the final result by merging all the rule sets generated for various clusters. 
Initially, we devised an algorithm that generates rule-sets, cluster by cluster, using the original FOLD algorithm of Shakerin et al. The problem with such an approach is that when the algorithm is focused on one cluster, it completely ignores the other surrounding clusters. Once a rule set has been induced that covers that cluster, the algorithm eliminates the covered samples in that cluster.
The elimination strategy for covered samples during the learning process greatly impacts the quality of literal selection. \textit{Reducing} the weights of covered samples while calculating the information gain, rather than removing them altogether, is another way to deal with the covered samples and a good way to have them influence the selection of future literals that cover other clusters. Reduction of weight of covered positive samples in subsequent iteration, rather than their complete removal, is referred to as \textit{demotion} by us. Demoting the covered positive examples clusters helps us in reaching the global optimum while avoiding being stuck in the local optima.

This paper makes the following novel contributions: it proposes an algorithm called Kmeans-FOLD that combines clustering algorithm and a demotion strategy with FOLD algorithm for binary classification tasks from any tabular dataset. To the best of our knowledge, this is the first time clustering and demotion strategies have been employed in inductive logic programming to learn rules. The experiments on UCI benchmark datasets suggest our new algorithm generates a more concise set of rules while maintaining the same accuracy as the FOLD algorithm. Thus, the Kmeans-FOLD algorithm has better explainability. Often, it also has higher accuracy. The Kmeans-FOLD can also output natural language explanation for the generated rule set as long as natural language descriptions of various literals are given. 

\section{ILP Problem Definition}
\label{sec:background}
Inductive Logic Programming (ILP) \cite{ilp} is a subfield of machine learning that learns models in the form of logic programming rules (Horn Clauses) that are comprehensible to humans. This problem is formally defined as:\\
\textbf{Given}
\begin{enumerate}
    \item a background theory $B$, in the form of an extended logic program, i.e., clauses of the form $h \leftarrow l_1, ... , l_m,\ not \ l_{m+1},...,\ not \ l_n$, where $l_1,...,l_n$ are positive literals and \textit{not} denotes \textit{negation-as-failure} (NAF) and $B$ has no even cycle
    \item two disjoint sets of ground target predicates $E^+, E^-$ known as positive and negative examples respectively
    \item a hypothesis language of function free predicates $L$, and a  refinement operator $\rho$ under $\theta-subsumption$ \cite{plotkin70} that would disallow even cycles.
\end{enumerate}
\textbf{Find} a set of clauses $H$ such that:
\begin{itemize}
    \item $ \forall e \in \ E^+ ,\  B \cup H \models e$
    \item $ \forall e \in \ E^- ,\  B \cup H \not \models e$
    \item $B \land H$ is consistent.
\end{itemize}

\section{Learning Default Theories}
\label{sec:Fold}
ILP algorithms induce logic programs that contain negated goals of the form \textit{not p}, where the \textit{not} is considered as \textit{negation as failure}. Logic programs containing negation-as-failure (with nonmonotonic semantics based on stable models \cite{cacm-asp}) are more expressive and concise when it comes to describing concepts (e.g., default reasoning, used by humans in common sense reasoning) \cite{cacm-asp,gelfond-kahl}. The extension of logic programming with negation-as-failure (under the stable model semantics) is termed answer set programming (ASP). Given that ASP can be used to model the human thought process, it is an ideal formalism to capture the logic encapsulated in a model that underlies a dataset.

The Kmeans-FOLD algorithm is an extension of the FOLD algorithm \cite{fold}. Kmeans-FOLD induces nonmonotonic logic programs that are more precise and more concise compared to the FOLD algorithm. Therefore, we first briefly discuss the FOLD algorithm. The FOLD algorithm learns a concept in terms of a default and possibly multiple exceptions (and exceptions to exceptions, exceptions to exceptions of exceptions, and so on). 


In the FOLD algorithm, each literal selection process must rule out some already covered negative examples without decreasing the number of positive examples covered. Significantly, no negative literal is used at this stage. Once the heuristic score (i.e., \textit{information gain}) (IG) becomes zero, or the maximum clause length is reached (whichever happens first), this process stops. At this point, if any negative example is still covered, they must be either noisy data or 
exceptions to the current hypothesis. Exceptions could be learned by swapping the current positive and negative examples, then calling the same algorithm recursively. As a result of this recursive process, FOLD can learn exceptions, exceptions to exceptions, and so on. In presence of noise, FOLD identifies and \textit{enumerates} noisy samples, that is, outputs them as ground facts in the hypothesis, to make sure that the algorithm converges. \textit{Maximum Description Length Principle} \cite{foil} is incorporated to heuristically control the hypothesis length and identify noise. Algorithm \ref{algo:fold} presents FOLD's pseudo-code.
\begin{algorithm}[!h]
\caption{FOLD Algorithm}
\label{algo:fold}
\begin{algorithmic}[1]
\Input $target,B,E^+,E^-$ 
\Output  $R = \{r_1,...,r_n\}$ \Comment{rule set}
\Function{FOLD}{$E^+,E^-,F_{used}$} 
\State $R \gets \emptyset$
\While{$|E^+| > 0$}
\State $\hat{r} \gets$ \Call{learn\_rule}{{$E^+$},{$E^-$},{$F_{used}$}}
\State $E^+ \gets E^+ \setminus covers(\hat{r},E^+,True)$
\State $R \gets R \cup \{ \hat{r} \}$
\EndWhile
\State \Return $R$
\EndFunction
\Function{LEARN\_RULE}{${E^+},{E^-},{F_{used}}$}
\While{$true$}
\State  $\hat{f} \gets$ \Call{add\_best\_literal}{{$E^+$},{{$E^-$}},{$F_{used}$}}
\State $F \gets F \cup \{ \hat{f} \}$
\State $\hat{r} \gets set\_default(\hat{r},F)$
\State $E^+ \gets covers(\hat{r},E^+,true)$
\State $E^- \gets E^- \setminus covers(\hat{r},E^-,false)$
\If{$\hat{f}$ is invalid or $|E^-| == 0$}
\If{$\hat{f}$ is invalid}
\State $F \gets F \setminus \{ \hat{f} \}$
\State $\hat{r} \gets set\_default(\hat{r},F)$
\Else 
\State $flag \gets true$
\EndIf
\State break
\EndIf
\EndWhile
\If{$flag$}
\State $AB \gets$ \Call{fold}{{$E^-$},{{$E^+$}},{$F_{used} + F$}}
\State $\hat{r} \gets set\_exception(\hat{r},AB)$
\EndIf
\State \Return $\hat{r}$
\EndFunction

\end{algorithmic}
\end{algorithm}

\begin{exmp}
\label{ex:pinguin}
$ B, E^+, E^-$ are background knowledge, positive and negative examples, respectively. The target, i.e., the predicate being learned is \texttt{fly(X)}.
\end{exmp}
\begin{verbatim}
B:  bird(X) :- penguin(X).
    bird(tweety).   bird(et).
    cat(kitty).     penguin(polly).
E+: fly(tweety).    fly(et).
E-: fly(kitty).     fly(polly).
\end{verbatim}

By calling FOLD, at line 3 while loop, the clause 
\texttt{\{fly(X) :- true.\}} is specialized. Inside the $LEARN\-RULE$ function, 
at line 10, the literal \texttt{bird(X)} is selected to add to the current clause, to get the 
clause $\hat{r}$ = \texttt{fly(X) :- bird(X)}, which happens to have the greatest IG among \texttt{\{bird,penguin,cat\}}. Then, at lines 15-16 the following updates 
are 
performed: $E^+=\{\}$,\ $E^-=\{polly\}$. A negative example, 
$polly$, a penguin, is still covered. In the next iteration, $LEARN\-RULE$ fails 
to introduce a positive literal to rule it out since the best IG in this case 
is zero. Therefore, the FOLD function is called recursively by swapping the 
$E^+$ and $E^-$ to learn exceptions with parameters $E^+ = \{polly\}$, $E^-=\{\}$. 
The recursive call (line 28), returns \texttt{\{fly(X) :- penguin(X)\}} but it's rejected because of lower information gain. In line 
28, a new predicate \texttt{ab0} is introduced and at lines 28-29 the clause 
\texttt{\{ab0(X) :- penguin(X)\}} is learned and added to the set of discovered 
abnormalities, namely, AB. In line 29, the negated exception (i.e \texttt{not 
ab0(X)}) and the default rule's body (i.e \texttt{bird(X)}) are compiled 
together to form the following theory:
\begin{center}
    \begin{tabular}{l}
        \texttt{fly(X) :- bird(X), not ab0(X).}\\     
        \texttt{ab0(X) :- penguin(X).}     
    \end{tabular}
\end{center}

Once the FOLD algorithm terminates and a hypothesis is created, it would iterate through each clause's body and eliminate the redundant or counterproductive predicates. These are the predicates whose elimination does not make the clause cover significant number of negative examples. Next, FOLD sorts the hypothesis clauses in ascending order based on the number of positive examples each clause covers. Then, starting from the smallest, FOLD eliminates each clause and measures the coverage of positive examples. If elimination of a clause does not affect the overall coverage, the clause is removed from the hypothesis permanently.

\subsection{Kmeans++ Algorithm}
Clustering is a classic technique in data mining, which aims to partition a group of unlabeled data samples into clusters. K-means algorithm is a popular solution for clustering by finding the cluster centers while minimizing the within-class variances with the given parameter $k$ as the number of final clusters. 
The K-means algorithm can be described as follows:
(i) sample $k$ points within the hypothesis space of the dataset as the initial cluster centers,
(ii) assign each data sample to its nearest cluster,
 (iii) calculate the centroid of data sample in each cluster as a new cluster center,
(iv) repeat steps (ii) and (iii) till the new clusters do not move.
However, the K-means algorithm suffers from two problems:
\begin{itemize}
\item the time complexity is a super-polynomial in the size of input dataset \cite{kmeans}.
\item the quality of the result relies on the quality of initial centers selection, and to find optimal initial points is an NP-hard problem.
\end{itemize}
The Kmeans++ \cite{kmeans++} aims to solve the second problem by using a novel method of sampling initial cluster centers. The sampling method can be described as:
(i) uniformly sample the first cluster center in the hypothesis space.
(ii) sample a point from the input data samples with the probability proportional to their squared distance to their nearest existing cluster center.
(iii) repeat step (ii) till $k$ cluster centers have been generated.

The output of ILP learning is a logic program rule set, in which, each rule covers a part of positive samples in the hypothesis space. Ideally, the rule set covers all the positive samples and avoids covering any negative samples, while generating as few rules as possible. Fewer rules means that they are easier for a human beings to understand. It also means less computational workload for the computer while running the algorithm. Most of mid-size or large dataset would likely have positive or negative examples scattered in several clusters in hypothesis space. Logic programming rules are well suited for this scenario where each positive example cluster is covered by a subset of rules.  
Literal selection method in a rule learning algorithm is very important as it will impact how many positive clusters are found in the hypothesis space. The literal selection process can possibly split up a positive cluster into several smaller clusters because only information gain is considered in the selection process. It is possible the method can erroneously select a few positive samples from one cluster and few more positive samples from another cluster to create a new cluster which will then be used to induce rules. If clusters are split up in this way, the rule set generated will have more rules to cover as our selection method leads to an increase in the number of clusters. An erroneous selection that splits up clusters amounts to getting stuck in local optima.

Our idea in this paper is to cluster positive samples \textit{before} the training phase, then apply ILP learning on each positive cluster along with the negative data samples that are present. Clustering in advance ensures that clusters of positive samples are not unnecessarily broken up leading to spurious rules being learned.

\subsection{Demotion strategy}

In the learning process of the FOLD algorithm, each literal selection step 
would rule out false negative samples implied by the new selected literal, and each logic rule learning step
would rule out true positive samples implied by the new rule. The literal selection process cuts the hypothesis space. With the elimination of training samples, once they have been used for generating rules, the region of focus shrinks. However, there is another way to deal with these already covered data samples, namely, reducing the weights of these samples rather than removing them. In effect, we are giving these samples lower weight in the next iteration of the algorithm rather than removing them altogether. We refer to this reduction of weight as demotion. Our demotion strategy allows us to avoid getting stuck in local optima. 

As stated above, the novel idea in this paper is to apply ILP learning on each positive sample cluster. When the learning process starts from a specific positive cluster, it will not take into account the surrounding clusters if those clusters' samples have not been included. The demotion strategy will slightly sacrifice  the quality of local literal selection, but it allows the surrounding clusters to contribute to selection of the literal. We demote the covered samples by reducing their weight rather than removing them altogether from future consideration.  Demoted samples will contribute to the information gain.
The heuristic for a given clause is calculated as follows:

\begin{equation}
\begin{aligned}
IG(L,R) = & t\left(log_2 \frac{p_1}{p_1 + n_1} - log_2 \frac{p_0}{p_0+ n_0} \right) \\
+ & t' \cdot f \left(log_2 \frac{p_3 \cdot f}{p_3 + n_3} - log_2 \frac{p_2 \cdot f}{p_2 + n_2} \right)
\end{aligned}
\end{equation}

\noindent where $L$ is the candidate literal to add to rule $R$, $p_0$ (resp. $p_2$) is the number of positive instances (resp. demoted instances) implied by the rule $R$, $n_0$ (resp. $n_2$) is the number of negative instances (resp. demoted instances) implied by the rule $R$, $p_1$ (resp. $p_3$) is the number of positive instances (resp. demoted instances) implied by the rule $R+L$, $n_1$ (resp. $n_3$) is the number of negative instances (resp. demoted instances)
implied by the rule $R+L$, $t$ (resp. $t'$)  is the number of positive instances (resp. demoted instances) implied by $R$ also covered by $R+L$, and $f$ is the demoting weight factor.

The demotion strategy avoids local optima and helps us find the global optima.
We get stuck in a local optima in situations where we select a literal with high information gain that will lead us down a path where subsequent literals added will produce low information gain. In contrast, we could have chosen a literal with low information gain, but one which leads to subsequent literals that produce very high information gain. The demotion strategy selects literals in a clause that combined together produce better information gain, leading us potentially to the global optimum.

\subsection{Kmeans-FOLD Algorithm}
\label{sec:fold+lime}
In this section, we introduce the Kmeans-FOLD algorithm by integrating Kmeans++, FOLD, and demotion strategy. This yields a powerful ILP algorithm capable of learning very concise logic programs from a transformed dataset. The new algorithm outperforms FOLD and ALEPH \cite{aleph} which are state-of-the-art ILP systems. 
As stated above, the FOLD algorithm divides the hypothesis space without considering the distribution of positive samples. This results in clusters of positive samples not being optimally discovered (Fig. \ref{fig:literalselection}), leading to extra rules being unnecessarily learned. Clustering the positive samples before the FOLD algorithm is invoked, avoids this problem of splintering positive clusters. However, this leads to another problem, namely, that our selection process for literals becomes too focused on a single cluster and becomes oblivious of other positive clusters. Keeping the positive clusters for which a literal has been selected, biases the next literal towards one that has some (conjunctive) relationship with the previous (demoted) cluster rather than selecting a completely independent literal. Thus, a combination of clustering and demotion leads to a logic program being learned that captures the model underlying the data more closely. This program is also very concise.

Algorithm \ref{algo:kmeansfold} shows how the Kmeans-FOLD processes a tabular dataset, where $E^+$ represents positive examples and $E^-$ negative ones. The FOLD algorithm only takes binary tabular data as input. Therefore, the input data also needs to be discretized. We use only one-hot encoding with range merging, since the C4.5 encoding would reduce the dimension based on information gain while disrupting the original distribution of input data. The algorithm first clusters the positive dataset with Kmeans++ algorithm into $k$ clusters, labeling data sample with the set of labels $L$. 
Then, after extracting the samples of each cluster with $SELECT$ function, the FOLD algorithm with demotion strategy ($D\_FOLD$ function) is applied to each positive cluster while keeping the entire negative samples $E^-$. The $D\_FOLD$ function takes the following parameters: positive examples, negative examples, demotion weight factor, demoted positive examples, and demoted negative examples. The generated rule set $R_i$ for each positive cluster $E^+_i$ would cover most of positive samples within its cluster. The merged rule set $R$ is the final result for the entire input dataset. The FOLD algorithm has been extended to handle numeric features (the extended algorithm is called FOLD-R). We have also incorporated the Kmeans++ clustering and demotion strategy in the FOLD-R algorithm. 

\begin{algorithm}
\caption{Kmeans-FOLD}
\label{algo:kmeansfold}
\begin{algorithmic}[1]
\Input $k:$ number of positive clusters
\Input $f:$ factor for sample weight demotion
\Input $target,B,E^+,E^-$ 
\Output  $R = \{r_1,...,r_n\}$ \Comment{rule set}

\Function{Kmeans\_FOLD}{$E^+,E^-,k,f$} 
\State $R \gets \emptyset$
\State $L \gets$ \Call{Kmeans++}{{$E^+$},{$k$}} \Comment{L: label for each sample}
\For{$i \gets 1$ to $k$}
\State $E^+_i \gets$ \Call{select}{{$E^+$},{$L$},{$i$}}
\State $R_i \gets$ \Call{D\_FOLD}{{$E^+_i$},{$E^-$},{$f$},{$E^+ \setminus E^+_i$}, {$\emptyset$}}
\State $R \gets R \cup R_i$
\EndFor
\State \Return $R$
\EndFunction
\end{algorithmic}
\end{algorithm}

\section{Experiments and Performance Evaluation}
\label{sec:Experiments}
In this section, we present our experiments on standard UCI  benchmarks \cite{uci}. The ALEPH system \cite{aleph} is used as the baseline. ALEPH is a state-of-the-art ILP system that has been widely used in prior work. To find a rule, ALEPH starts by building the most specific clause, which is called the ``bottom clause",
that entails a seed example. Then, it uses a branch-and-bound algorithm to perform a general-to-specific heuristic search for a subset of literals from the bottom clause to form a more general rule. We set ALEPH to use the heuristic enumeration strategy, and the maximum number of branch nodes to be explored in a branch-and-bound search to 500K. We use the standard metrics including precision, recall, accuracy, and $F_1$ score to measure the quality of the results. 
\begin{figure}
    \includegraphics[width=0.48\textwidth]{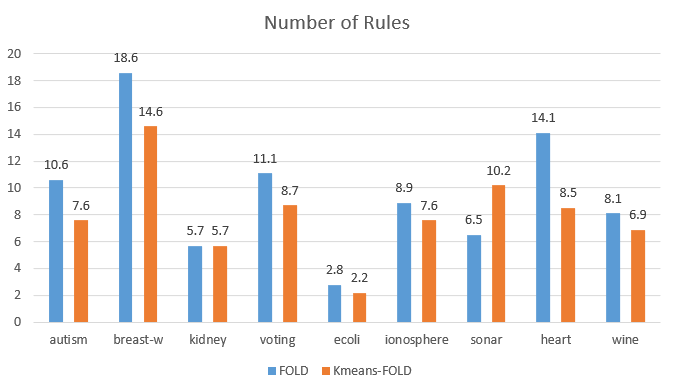}
\caption{Average Number of Rules Induced by Each Different Experiment}
\label{fig:numberofrules}
\end{figure}

We conducted two sets of experiments as follows: First, we run FOLD and Kmeans-FOLD algorithms to compare the average number of generated rules. The Kmeans-FOLD algorithm has a hyper-parameter $k$, representing the number of positive clusters, that ranges from 1 to 10. We run Kmeans-FOLD for each $k$. We select the rule set output of Kmeans-FOLD run that has accuracy similar to FOLD, picking one with the minimum number of rules. 
Second, we run ALEPH, FOLD, Kmeans-FOLD, FOLD-R, and Kmeans-FOLDR on 9 different datasets using the same 5-fold cross-validation setting. The $k$ in Kmeans-FOLD and Kmeans-FOLDR also range from 1 to 10, but we run twice for each $k$ for randomization of clustering. We choose the rule set with the best $F_1$ score as the result.


Figure \ref{fig:numberofrules} compares the average number of clauses generated by FOLD and Kmeans-FOLD on 9 UCI datasets. The FOLD and Kmeans-FOLD algorithm were both run 10 times on each of the 9 UCI datasets. The number of rules generated are reported as the average across the 10 runs. This is the reason why the number of rules generated is not a whole number.  With the exception of ``kidney" and ``sonar'', in all other datasets, Kmeans-FOLD discovers fewer number of rules. However, the accuracy of Kmeans-FOLD is much higher than the FOLD algorithm in ``sonar'' and is roughly the same in ``kidney".

\begin{table*}[h!]
\centering
\begin{tabular}{l|c|c|c|c|c|c|c|c|c|c|c|c|}
\cline{2-13}
                                 & \multicolumn{12}{c|}{Algorithm}                                                                                                                                                                                                                                                                                                                        \\ \hline
\multicolumn{1}{|l|}{Data Set}   & \multicolumn{4}{c|}{Aleph}                                                                                       & \multicolumn{4}{c|}{FOLD}                                                                                        & \multicolumn{4}{c|}{\textbf{Kmeans-FOLD}}                                                                        \\ \hline
\multicolumn{1}{|l|}{}           & \multicolumn{1}{l|}{Prec.} & \multicolumn{1}{l|}{Recall} & \multicolumn{1}{l|}{Acc.} & \multicolumn{1}{l|}{F1}   & \multicolumn{1}{l|}{Prec.} & \multicolumn{1}{l|}{Recall} & \multicolumn{1}{l|}{Acc.} & \multicolumn{1}{l|}{F1}   & \multicolumn{1}{l|}{Prec.} & \multicolumn{1}{l|}{Recall} & \multicolumn{1}{l|}{Acc.} & \multicolumn{1}{l|}{F1}   \\ \hline
\multicolumn{1}{|l|}{breast-w}   & 0.92                       & 0.87                        & 0.93                      & 0.89                      & 0.91                       & 0.86                        & 0.92                      & 0.88                      & 0.95                       & \textbf{0.98}               & \textbf{0.97}             & \textbf{0.97}             \\ \hline
\multicolumn{1}{|l|}{ecoli}      & 0.85                       & 0.75                        & 0.84                      & 0.80                      & 0.97                       & 0.42                        & 0.74                      & 0.58                      & 0.95                       & 0.55                        & 0.80                      & 0.70                      \\ \hline
\multicolumn{1}{|l|}{kidney}     & 0.96                       & 0.92                        & 0.93                      & 0.94                      & 0.99                       & 0.99                        & 0.99                      & 0.99                      & \textbf{1}                 & \textbf{1}                  & \textbf{1}                & \textbf{1}                \\ \hline
\multicolumn{1}{|l|}{voting}     & \textbf{0.97}              & 0.94                        & 0.95                      & 0.95                      & 0.92                       & 0.92                        & 0.94                      & 0.92                      & \textbf{0.95}              & 0.96                        & \textbf{0.97}             & \textbf{0.96}             \\ \hline
\multicolumn{1}{|l|}{autism}     & 0.73                       & 0.43                        & 0.79                      & 0.53                      & 0.89                       & 0.81                        & 0.92                      & 0.84                      & 0.87                       & 0.82                        & 0.92                      & 0.84                      \\ \hline
\multicolumn{1}{|l|}{ionosphere} & 0.89                       & 0.87                        & 0.85                      & 0.88                      & \textbf{0.96}              & 0.78                        & 0.91                      & 0.86                      & 0.95                       & 0.85                        & 0.93                      & 0.90                      \\ \hline
\multicolumn{1}{|l|}{sonar}      & 0.74                       & 0.56                        & 0.66                      & 0.64                      & 0.83                       & 0.66                        & 0.75                      & 0.73                      & \textbf{0.91}              & \textbf{0.93}               & \textbf{0.91}             & \textbf{0.92}             \\ \hline
\multicolumn{1}{|l|}{heart}      & 0.76                       & 0.75                        & 0.78                      & 0.75                      & 0.71                       & 0.62                        & 0.71                      & 0.66                      & 0.83                       & 0.70                        & 0.80                      & 0.75                      \\ \hline
\multicolumn{1}{|l|}{wine}       & 0.94                       & 0.86                        & 0.93                      & 0.89                      & 0.88                       & 0.83                        & 0.92                      & 0.85                      & 0.85                       & 0.88                        & 0.93                      & 0.86                      \\ \hline \hline
\multicolumn{1}{|l|}{Average}    & 0.86                       & 0.77                        & 0.85                      & 0.81                      & 0.90                       & 0.76                        & 0.87                      & 0.81                      & 0.92                       & 0.85                        & 0.91                      & 0.88                      \\ \hline
\end{tabular}
\caption{Evaluation of ALEPH, FOLD, and Kmeans-FOLD with 9 UCI Datasets}
\label{tbl:accuracies}
\end{table*}

\begin{table*}[]
\centering
\begin{tabular}{l|c|c|c|c|c|c|c|c|c|c|c|c|}
\cline{2-13}
                                 & \multicolumn{12}{c|}{Algorithm}                                                                                                                                                                                                                                                                                                                        \\ \hline
\multicolumn{1}{|l|}{Data Set}   & \multicolumn{4}{c|}{Aleph}                                                                                       & \multicolumn{4}{c|}{FOLD-R}                                                                                      & \multicolumn{4}{c|}{\textbf{Kmeans-FOLDR}}                                                                       \\ \hline
\multicolumn{1}{|l|}{}           & \multicolumn{1}{l|}{Prec.} & \multicolumn{1}{l|}{Recall} & \multicolumn{1}{l|}{Acc.} & \multicolumn{1}{l|}{F1}   & \multicolumn{1}{l|}{Prec.} & \multicolumn{1}{l|}{Recall} & \multicolumn{1}{l|}{Acc.} & \multicolumn{1}{l|}{F1}   & \multicolumn{1}{l|}{Prec.} & \multicolumn{1}{l|}{Recall} & \multicolumn{1}{l|}{Acc.} & \multicolumn{1}{l|}{F1}   \\ \hline
\multicolumn{1}{|l|}{breast-w}   & 0.92                       & 0.87                        & 0.93                      & 0.89                      & 0.93                       & 0.91                        & 0.95                      & 0.92                      & \textbf{0.96}              & 0.97                        & \textbf{0.97}             & 0.96                      \\ \hline
\multicolumn{1}{|l|}{ecoli}      & 0.85                       & 0.75                        & 0.84                      & 0.80                      & 0.93                       & 0.87                        & 0.92                      & 0.90                      & \textbf{0.98}              & \textbf{0.96}               & \textbf{0.98}             & \textbf{0.97}             \\ \hline
\multicolumn{1}{|l|}{kidney}     & 0.96                       & 0.92                        & 0.93                      & 0.94                      & \textbf{1}                 & 0.97                        & 0.98                      & 0.99                      & \textbf{1}                 & \textbf{1}                  & \textbf{1}                & \textbf{1}                \\ \hline
\multicolumn{1}{|l|}{voting}     & \textbf{0.97}              & 0.94                        & 0.95                      & 0.95                      & 0.92                       & 0.95                        & 0.95                      & 0.94                      & 0.93                       & \textbf{0.98}               & 0.96                      & 0.95                      \\ \hline
\multicolumn{1}{|l|}{autism}     & 0.73                       & 0.43                        & 0.79                      & 0.53                      & 0.83                       & 0.88                        & 0.92                      & 0.86                      & \textbf{0.91}              & \textbf{0.96}               & \textbf{0.96}             & \textbf{0.94}             \\ \hline
\multicolumn{1}{|l|}{ionosphere} & 0.89                       & 0.87                        & 0.85                      & 0.88                      & 0.90                       & 0.92                        & 0.88                      & 0.91                      & 0.92                       & \textbf{0.98}               & \textbf{0.93}             & \textbf{0.95}             \\ \hline
\multicolumn{1}{|l|}{sonar}      & 0.74                       & 0.56                        & 0.66                      & 0.64                      & 0.72                       & 0.76                        & 0.71                      & 0.72                      & 0.87                       & 0.86                        & 0.85                      & 0.86                      \\ \hline
\multicolumn{1}{|l|}{heart}      & 0.76                       & 0.75                        & 0.78                      & 0.75                      & 0.73                       & 0.76                        & 0.77                      & 0.74                      & \textbf{0.84}              & \textbf{0.81}               & \textbf{0.85}             & \textbf{0.82}             \\ \hline
\multicolumn{1}{|l|}{wine}       & 0.94                       & 0.86                        & 0.93                      & 0.89                      & 0.94                       & 0.94                        & 0.97                      & 0.94                      & \textbf{1}                 & \textbf{1}                  & \textbf{1}                & \textbf{1}                \\ \hline \hline
\multicolumn{1}{|l|}{Average}    & 0.86                       & 0.77                        & 0.85                      & 0.81                      & 0.88                       & 0.89                        & 0.89                      & 0.88                      & \textbf{0.93}              & \textbf{0.95}               & \textbf{0.95}             & \textbf{0.94}             \\ \hline
\end{tabular}
\caption{Evaluation of ALEPH, FOLD-R, and Kmeans-FOLDR with 9 UCI Datasets}
\label{tbl:accuracies2}
\end{table*}

Table \ref{tbl:accuracies} and \ref{tbl:accuracies2} presents the comparison of classification metrics on each of the 9 UCI datasets. The best performer is highlighted with boldface font. The Kmeans-FOLDR produces a classifier with the best $F_1$ score in 5 cases, the Kmeans-FOLD produces the best results in the reamaining 4 cases. Both Kmeans-FOLD and Kmeans-FOLDR have better overall scores compared to their predecessors (FOLD and ALEPH). It is worth noting that all the FOLD series methods have better $F_1$ scores than ALEPH, which shows that answer set programs can describe a model better than standard logic programs without negation. All experiments were run on an Intel Core i5-10400 @ 2.9GHz with 32 GB RAM and a 64-bit Windows 10. The FOLD, FOLD-R, Kmeans-FOLD, and Kmeans-FOLDR are implemented with Python3. ALEPHv.5 has been ported into SWI-Prolog by \cite{alephswiprolog}.



\section{Output in a Natural Language}

The output of the FOLD family of algorithms is an answer set program. Each ASP rule can be understood on its own and can be translated into natural language, as long as we have natural language rendering of each of the predicates involved. To make the outpur more human readable, the Kmeans-FOLD and Kmeans-FOLDR algorithm's output is translated into natural language (English). The following logic program is induced by running Kmeans-FOLDR algorithm on Breast Cancer Wisconsin dataset:

{\scriptsize 
\begin{verbatim} 
mt(X) :-bare_nuclei(X, N1), N1 > 9, N1 =< 10,
        bland_chromatin(X, N2), N2 > 7, N2 =< 8.
mt(X) :- single_epi_cell_size(X, N3), N3 > 6, N3 =< 7.
mt(X) :- bare_nuclei(X, N1), N1 > 9, N1 =< 10., 
mt(X) :- bland_chromatin(X, N2), N2 > 7, N2 =< 8. 
mt(X) :- clump_thickness(X, N6), N6 > 9.
mt(X) :- clump_thickness(X, N6), N6 > 3, N6 =< 4, 
         not ab1(X), not ab2(X). 
ab1(X) :- cell_shape_uniformity(X, N3), N3 > 1, N3 =< 2,
          normal_nucleoli(X, N4), N4 > 1, N4 =< 2. 
ab2(X) :- cell_shape_uniformity(X, N3), N3 > 1, N3 =< 2,
          marginal_adhesion(X, N5), N5 =< 1,
          bare_nuclei(X, N1), N1 =< 1,
          bland_chromatin(X, N2),N2 > 2, N2 =< 3. 
\end{verbatim}
}

\noindent 
Next, we will have to provide the English description of the predicate. This is declared using the \#pred directive, for example:

\smallskip
\noindent
{\scriptsize 
{\tt \#pred cell\_shape\_uniformity(X, N): cell shape uniformity of @X is @N}
}

\smallskip
\noindent 
{\scriptsize 
{\tt \#pred mt(X): Tumor @X is malignant}
}

\noindent For abnormal predicates (exceptions), our algorithm has a fixed numbering scheme, and so the translation algorithm will translate {\tt ab$_I$} as:   

\smallskip
\noindent
{\scriptsize 
{\tt \#pred ab$_I$(X): abnormal condition I applies to tumor X}
}

\smallskip\noindent 
With these declarations that provide the natural language reading of each predicate, we obtain the following natural language translation of the last three rules above: 

{\scriptsize 
\begin{verbatim}
tumor X is malignant if:
    the clump thickness of X is larger than 3 and less 
        than or equal to 4
    unless abnormal condition 1 applies and 
           abnormal condition 2 applies.
abnormal condition 1 applies to tumor X if:
    cell shape uniformity of X is larger than 1 and 
        less than or equal to 2,
    normal nucleoli level of X is larger than 1 and 
        less than or equal to 2.
abnormal condition 2 applies to tumor X if:
    cell shape uniformity of X is less than or equal to 2,
    marginal adhesion level of tumor X is less than 1,
    bare nuclei reading of tumor X is less than, and
    bland chormatin level of tumor X is more than 2 
        but less than equal to 3.
\end{verbatim}  
}

\section{Related Work}
\label{sec:related}
A survey of ILP can be found in \cite{ilp20}. Rule extraction from statistical Machine Learning models has been a long-standing goal of the community. The rule extraction algorithms from machine learning models are classified into two categories: 1) Pedagogical (i.e., learning symbolic rules from black-box classifiers without opening them) 2) Decompositional (i.e., to open the classifier and look into the internals). TREPAN \cite{trepan} is a successful pedagogical algorithm that learns decision trees from neural networks. SVM+Prototypes \cite{svmplus} is a decompositional rule extraction algorithm that makes use of KMeans clustering to extract rules from SVM classifiers by focusing on support vectors. Another rule extraction technique that is gaining attention recently is ``RuleFit" \cite{rulefit}. RuleFit learns a set of weighted rules from ensemble of shallow decision trees combined with original features. In ILP community also, researchers have tried to combine statistical methods with ILP techniques. Support Vector ILP \cite{svmilp} uses ILP hypotheses as kernel in dual form of the SVM algorithm. kFOIL \cite{kfoil} learns an incremental kernel for SVM algorithm using a FOIL style specialization. nFOIL \cite{nfoil} integrates the Naive-Bayes algorithm with FOIL. The advantage of our research over all of the above mentioned research work is that we generate answer set programs containing negation-as-failure that correspond closely the human thought process. Thus, the descriptions are more concise. Second it is scalable thanks to the greedy nature of our clause search. 

Shakerin et al. have also developed the SHAP-FOLD algorithm which makes use of Shapley values and High Utility Itemset Mining (HUIM) to guide the search in the FOLD algorithm \cite{shap-fold}. However, the SHAP-FOLD algorithm has high time complexity as generation of Shapley values and HUIM is NP hard, since combination of features have to be considered. Thus, SHAP-FOLD may not be highly scalable. In contrast, clustering algorithms are polynomial time, and therefore Kmeans-FOLD is more scalable than SHAP-FOLD. 


\section{Conclusions and Future Work}

In this paper, we presented a clustering based algorithm called Kmeans-FOLD for binary classification tasks. We also proposed the Kmeans-FOLDR algorithm that similarly extends the FOLD-R algorithm. FOLD-R algorithm is an extension of the FOLD algorithm to handle numeric features. Both Kmeans-FOLD and Kmeans-FOLDR algorithms learn very concise answer set programs. The learned answer set program can be translated into English to provide a natural language explanation. Our experiments on the UCI datasets suggest that the new Kmeans clustering and demotion-based algorithms can provide same or better performance as the original FOLD algorithm wrt accuracy, but output more concise answer set programs. 

There are many directions for our future exploration: (i) explore the performance of our algorithm with other clustering methods, such as spectral clustering, hierarchical clustering, and the K-means Elbow method. (ii) better methods for literal selection at early stages for dividing positive clusters;
(iii) better strategies for demotion: the demotion strategy we are using currently works well but is not optimal; (iv) new methods for multi-categorical classification, our algorithm has to be executed to build multiple models for different classes for multi-categorical classification. (v) extending our algorithm for multi-relational input datasets. Note that our work reported here that makes use of clustering and demotion to improve the FOLD algorithm is just the beginning. Many future improvements are possible: (i) right now we apply the clustering algorithm to the positive samples only once in the beginning; clustering after each iteration step will be interesting to study and may produce even better results, (ii) study if clustering negative examples also leads to improvement, (iii) develop methodology to set the hyper-parameters such as $k$ (the number of cluster) and $f$ (the demotion factor).


\bibliographystyle{aaai}
\bibliography{mycitations}

\end{document}